# Ambiguity in the Acquisition of Lexical Information[1]

**Lucy Vanderwende**

Microsoft Corporation
One Microsoft Way
Redmond, WA 98052-6399
lucyv@microsoft.com

## Abstract

This paper describes an approach to the automatic identification of lexical information in on-line dictionaries. This approach uses bootstrapping techniques, specifically so that ambiguity in the dictionary text can be treated properly. This approach consists of processing an on-line dictionary multiple times, each time refining the lexical information previously acquired and identifying new lexical information. The strength of this approach is that lexical information can be acquired from definitions which are syntactically ambiguous, given that information acquired during the first pass can be used to improve the syntactic analysis of definitions in subsequent passes. In the context of a lexical knowledge base, the types of lexical information that need to be represented cannot be viewed as a fixed set, but rather as a set that will change given the resources of the lexical knowledge base and the requirements of analysis systems which access it.

## Introduction

In order to identify lexical information automatically in an on-line dictionary, various techniques have been used, all of which employ defining formulae in some way. Defining formulae are 'significant recurring phrases' in dictionary definitions (Markowitz et al. 1986; p. 113). Some have implemented these defining formulae as string patterns that match the definition text, e.g., Chodorow et al. (1985) and Markowitz et al. (1986), while others have implemented them as structural patterns that match the syntactic analysis of the definition text, e.g., Jensen and Binot (1987), Alshawi (1989), Ravin (1990), Montemagni and Vanderwende (1992), and Klavans et al. (1990). The earlier studies used defining formulae as if there were a one-to-one relation between the formula and the type of lexical information it identified; for example, the defining formulae *used for* always identifies the Instrument type of lexical information.

Later studies (Ravin 1990 and Klavans et al. 1990) have shown that some defining formulae can convey several types of lexical information. The problem is that defining formulae sometimes must rely on ambiguous words in their patterns, such as *with*, *of* and *unit of* as shown below, and even in dictionaries which make use of a limited defining vocabulary, this ambiguity is not resolved. Ravin (1990) shows that *with* in the definitions of verbs can convey many types of semantic relations, e.g., the relation USE-OF-INSTRUMENT in the definition of *angle* (L 3,vi,1): 'to fish with a hook and line'[2], and MANNER in the definition of *attack* (L 1,v,4): 'to begin (something) with eagerness and great interest'. In Klavans et al. (1990), we see that the pattern *a unit of* can convey the relation AMOUNT/CURRENCY in the definition of *pice* (W n,1): 'a former monetary unit of India and Pakistan equal to ...', and it can convey SUBDIVISION in the definition of *division* (W n,3c1): 'the basic unit of men for administration ...'.

Both of these studies argue that a syntactic analysis of the dictionary definition is required in order to identify the lexical information reliably. Klavans et al. (1990) show how the relation conveyed by the pattern *a unit of* can be identified on the basis of syntactic information. Ravin (1990) describes a system for disambiguating the preposition *with* in verb definitions which requires a syntactic analysis of the definition, but also lexical information for the verb modified by *with* and the noun complement of *with*. For example, to determine that *with* conveys a USE-OF-INSTRUMENT relation in the definition of *angle*: 'to fish with a hook ...', lexical information is needed for both the verb *fish* and the noun *hook*. In order to acquire this lexical information, the definitions of *fish* and *hook* are parsed, after which heuristics, or patterns, are applied to the syntactic analysis to determine the necessary lexical information. The process described in Ravin (1990) can be seen as depth-first: in order to acquire the semantic relations for the definition of *angle*, at least some of the semantic relations for other words needs to be acquired first.

---

[1] Published in *Proceedings of the AAAI 1995 Spring Symposium Series*, working notes of the Symposium on Representation and Acquisition of Lexical Knowledge, pp. 174-179.

[2] The sense numbers are proceeded by the letter L, which indicates that the source of the definition is Longman Dictionary of Contemporary English.

In this paper, I will describe a different approach to solving this problem from that taken in Ravin (1990). This implementation involves multiple passes through the dictionary, and at each stage all the lexical information that can be reliably identified is added to the lexical knowledge base (LKB). We store the lexical information in the manner described in Dolan et al. (1993) and Richardson et al. (1993). This process can be seen as breadth-first: acquire lexical information for each word in the dictionary, then use that information to acquire more and/or more reliable lexical information. When all we are concerned with is to disambiguate the defining formula, the results of our approach will not differ from those of Ravin. However, when acquiring lexical information from definitions which are syntactically ambiguous, our approach shows better results.

## Processing the dictionary multiple times

During the first pass of our incremental approach, only the defining formulae which unambiguously identify lexical information will be used. Based on the syntactic analysis of a broad-coverage parser, such as PEG (Jensen 1986) used in Ravin's study and the Microsoft English Grammar used in this study, a number of semantic relations can be identified, e.g., HYPERNYM, INSTRUMENT-OF, MATERIAL, PART, and PART-OF. For some of these relations, not all of their possible patterns, or defining formulae, will be applied at this stage, e.g., for PART-OF, the unambiguous pattern *part of*[3] can be applied, but not the pattern *of*, which will be discussed later. The lexical information in (1) has been identified by applying the *part of* pattern to the syntactic analysis of the definition of *flower* (L 1,n,1): 'the part of a plant, often beautiful and colored, ... '

(1) [flower] → (PART-OF) → [plant]

Similarly, for the relation PART, the unambiguous pattern *{that,which} {has,have}* can be applied; the lexical information in (2) has been identified by applying this pattern to the syntactic analysis of the definition of *plant* (L 2,n,1): 'a living thing that has leaves and roots, and grows usu. in earth, ... '.

(2) [plant] → (PART) → [leaf, root]

Once the semantic relations acquired during the first pass have been added to the LKB, the lexical information exists which will allow the ambiguous words in the defining formulae to be disambiguated, enabling the identification of more semantic relations during subsequent passes. One of the patterns for identifying a PART-OF relation which could not be applied during the first pass is the pattern characterized by *of*. The preposition *of* can convey many different relations, among which are PART-OF, MATERIAL, and HYPERNYM. Examples of definitions in which *of* conveys PART-OF, MATERIAL, and HYPERNYM relations are (3)-(5), respectively.

---
[3]There are further restrictions to the pattern *part of*, namely that *part* should be identified as the HYPERNYM. For the sake of brevity, the patterns will be given only in abbreviated form in this paper. For a more detailed account of structural patterns, see Montemagni and Vanderwende (1992).

(3) *clove* (L 1,n): 'the dried unopened flower of a tropical Asian plant, used ... '

(4) *bullion* (L n): 'bars of gold or silver'

(5) *christening* (L,n): 'the Christian ceremony of baptism ...'

In order to determine the relation conveyed by *of*, it is necessary to access the lexical information of the noun modified by *of* and the complement of *of*. The pattern which identifies PART-OF from an *of* prepositional phrase (*of*-PP) can be paraphrased as:

(a) if the modified noun has a PART-OF relation whose value matches the *of*-complement, or
(b) if the complement has a PART relation whose value matches the modified noun,
then the *of*-PP conveys the relation PART-OF.

Consider the definition of *clove* (L 1,n): 'the dried unopened flower of a tropical Asian plant, used ... ' Condition (a) applies to this definition, because the modified noun, *flower*, has a PART-OF relation, as shown in (1), and its value, namely *plant*, is the same as the complement of the *of*-PP. The lexical information identified from this definition is shown in (6).

(6) [clove] → (PART-OF) → [plant]

The results of the incremental approach described until now do not differ from those of Ravin; both can handle ambiguous defining formulae well. The only difference is that in our approach, the lexical information for *flower* and *plant* will already be available when the entry for *clove* is being processed, while in Ravin's approach the definitions of *flower* and *plant* will need to be analyzed before the PART-OF relation can be identified in the definition of *clove*.

The incremental approach shows clear advantage, however, in acquiring lexical information from definitions which are syntactically ambiguous. The semantic relations that have been acquired on a first pass not only serve to disambiguate defining formulae, but they can also be used to improve the syntactic analysis of dictionary definitions in subsequent passes.

## Syntactic analysis during subsequent passes

The syntactic analysis to which the patterns are applied is the output of the Microsoft English Grammar, a broad-coverage grammar. Guided by rule and part-of-speech probabilities, the parsing algorithm produces the most probable parse first (see Richardson, 1994). The method for dealing with ambiguity at this initial, syntax-only, stage 'is to attach pre- and post-modifiers in a single arbitrary pattern (usually to the closest possible head, ...)' (Jensen and Binot 1987). In order to solve the ambiguities of the initial syntactic analysis, Jensen and Binot (1987) proposed to access the semantic information that can be identified automatically in an on-line dictionary. The goal of the work under discussion is to provide this level of semantic information, and the relations that can be identified in a definition during the first pass of the on-line dictionary fulfill this goal, but only when the definition is not syntactically ambiguous.

Most dictionary definitions, however, are syntactically ambiguous. All of the sources of ambiguity that occur in free text are also found in definition texts: prepositional phrases, relative clauses, and participle clauses can all be multiply attached. The most common source of ambiguity in definitions, however, is coordination. While the method of attaching post-modifiers to the closest possible head (i.e., right attachment) produces the correct parse in the vast majority of cases, this method is less successful in solving ambiguity in coordination. Consider the definition of *plantain* in (7):

(7) *plantain* (L n): 'a type of common wild plant with wide leaves growing close to the ground and small green flowers'

The placement of the constituent *small green flowers* is ambiguous; *small green flowers* can be coordinated with *ground* to form the constituent *the ground and small green flowers*, or it can be coordinated with *leaves* to form the constituent *wide leaves and small green flowers*. Because the initial syntactic analysis arbitrarily attaches an ambiguous constituent to its closest head, *small green flowers* forms a constituent with *ground*. This is illustrated in the syntactic analysis in figure 1, which has been excerpted to focus on the coordination ambiguity. Based on this analysis, the pattern for the PART relation will identify that the *plantain* has *leaves*, but it will fail to find that the *plantain* has *flowers*.

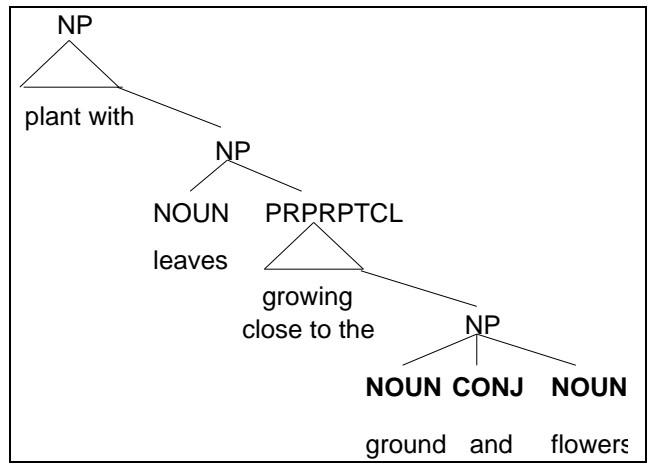

Figure 1. Schematic representation of the parse for the definition of *plantain* (L n) before reattachment

One of the heuristics for handling coordination is to check which terms are most similar. In this example, we should check whether *ground* and *flower* are more similar than *leaf* and *flower* are. To check the similarity, the lexical information acquired during the first pass from the definitions of *ground*, *leaf*, and *flower* will be compared. As we saw in (1) and (2), repeated here as (8) and (9), both *flower* and *leaf* are parts of a plant.

(8) *flower* (L 1,n,1): [flower] → (PART-OF) → [plant]

(9) *plant* (L 2,n,1): [plant] → (PART) → [leaf, root]

This similarity between *leaf* and *flower* ranks higher than the similarity found between *ground* and *flower*. The closest connection between *ground* and *flower* was found through the word *grow*, namely that flowers are grown (see *flower* (L 1,n,1) and that growing is located on the ground (see, e.g., *gourd* (L n,1)). Given the higher similarity between *leaf* and *flower*, the initial syntactic analysis is modified to reflect that *wide leaves and small green flowers* is a constituent; the revised analysis is shown in figure 2. Based on the revised syntactic analysis, the pattern for the PART relation will now identify that *plantain* has both *leaves* and *flowers*.

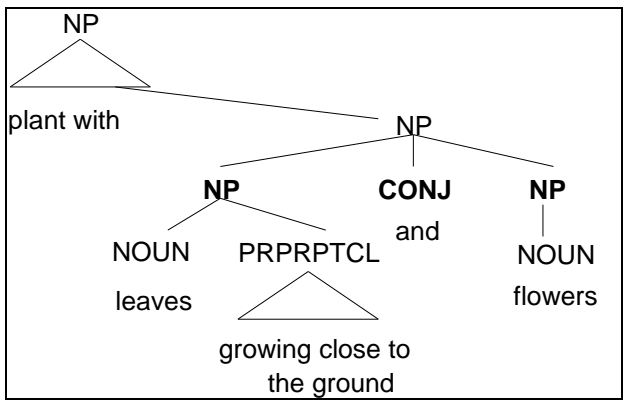

Figure 2. Schematic representation of the parse for the definition of *plantain* (L n) after reattachment

## Patterns and heuristics are merging

While the first pass identifies relations by using the initial syntactic analysis of the definitions, this example shows that more relations can be identified by disambiguating the initial syntactic sketch. And we saw that by disambiguating the defining formulae, more accurate relations can be identified. Sometimes disambiguating the initial syntactic sketch involves disambiguating (part of) the defining formulae, e.g., when determining the correct attachment of a PP. In this case, it is interesting to note that the patterns for identifying a particular relation are merging with, if not already quite the same as, the heuristics for determining the correct attachment.

Consider the definition of *angling* (L n): 'the sport of catching fish with a hook and line'. According to the strategy of right attachment, the initial syntactic analysis attaches the PP *with a hook and line* to the noun *fish*; the initial analysis is shown in figure 3. In order to determine the correct attachment of the PP, the system compares (a) *to catch with a hook and line* to (b) *a fish with a hook and line* (see Jensen and Binot (1987)). One of the reattachment heuristics checks for an INSTRUMENT relation between a head and the complement of *with*. During the initial pass, the INSTRUMENT relation in (10) was identified from the definition of *hook* (L 1,n,1): 'a curved piece of metal, plastic, etc., for catching something ... '.

(10) [hook] → (INSTRUMENT) → [catch]

Because there is an INSTRUMENT relation between the head *catch* and the complement *hook*, the analysis in (a) *to catch with a hook and line* is preferred, as shown in figure 4.

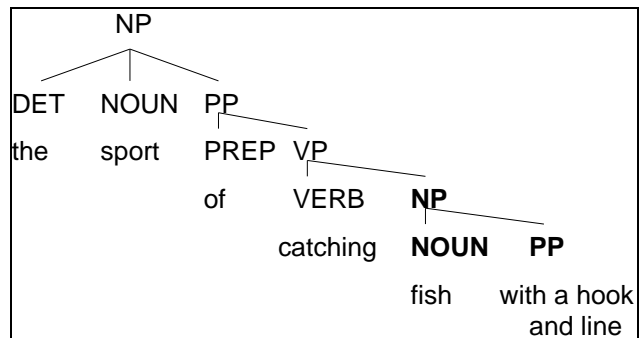

Figure 3. Schematic representation of the parse for the definition of *angling* (L n) before reattachment

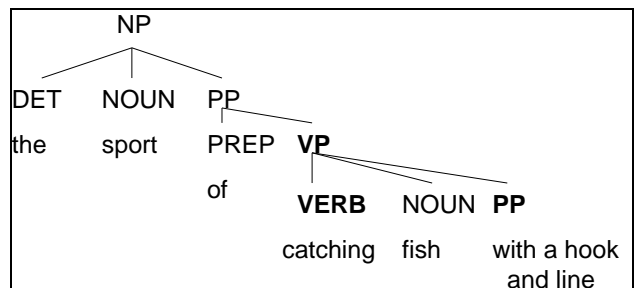

Figure 4. Schematic representation of the parse for the definition of *angling* (L n) after reattachment

One of the patterns that are applied to the syntactic analysis is the following INSTRUMENT pattern, which can be paraphrased as: if the *with*-PP modifies a verb, and if the complement of *with* has an INSTRUMENT relation whose value matches the modified verb, then identify an INSTRUMENT relation on the verb. We can see that the initial syntactic analysis of the definition of *angling* did not meet the conditions of this INSTRUMENT pattern because the *with*-PP modified the noun *fish*, and not the verb *catch*. The revised syntactic analysis, however, does meet the conditions and so the INSTRUMENT relation in (11) was identified[4]:

(11) [angling] → (INSTRUMENT) → [hook, line]

We can see that the heuristics for attaching a *with*-PP are quite similar to the pattern for identifying an INSTRUMENT relation from a *with*-PP. Both the task of determining the correct attachment and of determining the semantic relation need to access lexical information for the words in the text, whether free text or dictionary text.

---

[4]First, the relation [catch] → (INSTRUMENT) → [hook, line] is identified, but then also the relation in (11) [angling] → (INSTRUMENT) → [hook, line], because *catch*, as well as *sport*, is the HYPERNYM of *angling*.

The endeavor of discovering defining formulae and formulating their patterns has sometimes been criticized as "open-ended", due to the variation with which relations are specified in the dictionary (Veronis and Ide, 1993). While it is certainly the case that there are many ways in which a PART-OF or INSTRUMENT relation is conveyed, these patterns must be seen as a reusable tool, one which can also help solve ambiguity in free text. Moreover, these patterns can identify semantic relations in free text, as well as in dictionary text, and so they should be seen as part of a more general system for language understanding.

## Conclusion

The overall goal for acquiring lexical information automatically is to create an LKB that can be used in syntactic and semantic processing; in particular, lexical information that can motivate the correct analysis of PPs, relative clauses, and coordination. The task of processing dictionary definitions requires the correct analysis of the same phenomena just mentioned, as well as the disambiguation of defining formulae. By adopting an incremental approach to the acquisition of lexical information, we have seen that information extracted during the initial pass can be used to disambiguate the analyses and defining formulae during subsequent pass, enabling the acquisition of more, and more accurate, lexical information.

The approach is, therefore, not circular, a concern put forward by Veronis and Ide (1993); at no time is the ambiguity in one definition solved by chasing through the definitions of other definitions, a situation which quickly could lead to an infinite loop. The apparent circularity is managed by storing the results of each pass in the LKB for use during a subsequent pass over the dictionary, where the LKB improves in quality after each new pass.

An important result of implementing the patterns for identifying lexical information during subsequent passes is the discovery that they are similar to the heuristics for solving ambiguity problems in free text. In our case, we see these two sets of rules merging with no, or very few, differences. The dictionary patterns should be seen as a reusable tool and not of use only within the context of dictionary definitions.

Finally, taking an incremental approach to the acquisition of lexical information raises an important issue in the context of the representation of lexical information. As a system develops, different types of lexical information will need to be represented; at first, only a subset of the possible relations will be available. As the paper has shown, we can get better results from processing the dictionary automatically if we do not expect all of the lexical information to be there at once. A flexible representation schema is therefore necessary. Also, NLP is not at a stage where it is known a priori which semantic relations are necessary; at best, we currently have some studies of the semantic relations which are minimally required by a specific component (e.g., Vanderwende (1994) lists the relations necessary for a noun sequence analysis). We must first explore which semantic relations are needed, before trying to fix upon the ultimate representation of lexical information.

## Acknowledgments

I would like to thank Steve Richardson and George Heidorn for implementing the control structure which underlies the system described here; and Karen Jensen, Bill Dolan, Joseph Pentheroudakis and Diana Peterson for their feedback and support.## References

Alshawi, H. 1989. Analysing the Dictionary Definitions. In Boguraev & Briscoe, eds., Computational Lexicography for Natural Language Processing, Longman, London, pp. 153-170.

Chodorow, M.S., R.J. Byrd, G.E. Heidorn. 1985. Extracting semantic hierarchies from a large on-line dictionary. In Proceedings of the 23rd Annual Meeting of the ACL, pp. 299-304.

Dolan, W.B., L. Vanderwende and S.D. Richardson. 1993. Automatically deriving structured knowledge bases from on-line dictionaries. In Proceedings of the First Conference of the Pacific Association for Computational Linguistics, at Simon Fraser University, Vancouver, BC., pp.5-14.

Jensen, K. 1986. Parsing strategies in a broad-coverage grammar of English. Research Report RC 12147, IBM Thomas J. Watson Research, Yorktown Heights, NY.

Jensen, K., and J.-L. Binot. 1987. Disambiguating prepositional phrase attachments by using on-line dictionary definitions. In Computational Linguistics 13.3-4.pp. 251-60.

Klavans, J., M. Chodorow, and N. Wacholder. 1990. From Dictionary to Knowledge Base via Taxonomy. In Electronic Text Research, University of Waterloo, Centre for the New OED and Text Research, Waterloo, Canada.